\documentclass[letterpaper, 10pt, conference]{ieeeconf}  
\IEEEoverridecommandlockouts
\usepackage{cite}
\usepackage{amsmath,amssymb,amsfonts}
\usepackage{graphicx}
\usepackage{booktabs}
\usepackage{multirow}
\usepackage{array}
\usepackage{xcolor}
\usepackage{colortbl}
\usepackage{tabularx}
\definecolor{keptblue}{RGB}{226,236,248}
\usepackage{url}
\usepackage{eso-pic}   
\usepackage{textcomp}
\usepackage{cuted}      
\usepackage{capt-of}    
\usepackage{flushend}   
\usepackage[hidelinks,breaklinks=true,
  pdftitle={From Instrument-Mounted Demonstrations to In-Vivo Execution: Learning Bimanual Laparoscopic Appendectomy Without Robot-Collected Demonstrations},
  pdfauthor={Dongho Yee, Juahn Oh, Jinseok Lee, Jiyul Lee, Yechan Seo, Seong Jeong, Minsung Kim, Seonho Shim, Younghoon Noh, Hyuk Choi, Youngbin Kong, Kyu Eun Lee, Hyoun-Joong Kong},
  pdfsubject={Learning a bimanual laparoscopic appendectomy policy from instrument-mounted demonstrations and deploying it in vivo},
  pdfkeywords={surgical robotics, imitation learning, diffusion policy, laparoscopy, in-vivo deployment, instrument-state logger}
]{hyperref}
\graphicspath{{figures/}}
\makeatletter
\def\@IEEEtablestring{table}
\long\def\@makecaption#1#2{%
\ifx\@captype\@IEEEtablestring%
\noindent\parbox[t]{\hsize}{\footnotesize\raggedright #1:~#2}\par\@IEEEtablecaptionsepspace%
\else
\@IEEEfigurecaptionsepspace%
\setbox\@tempboxa\hbox{\footnotesize #1.~~ #2}%
\ifdim \wd\@tempboxa >\hsize%
\setbox\@tempboxa\hbox{\footnotesize #1.~~ }%
\parbox[t]{\hsize}{\footnotesize \noindent\unhbox\@tempboxa#2}%
\else%
\ifcenterfigcaptions \hbox to\hsize{\footnotesize\hfil\box\@tempboxa\hfil}%
\else \hbox to\hsize{\footnotesize\box\@tempboxa\hfil}%
\fi\fi\fi}
\makeatother

\title{\LARGE \bf From Instrument-Mounted Demonstrations to In-Vivo Execution:\\Learning Bimanual Laparoscopic Appendectomy Without Robot-Collected Demonstrations}

\author{Dongho Yee$^{*,1,2,4,5}$, Juahn Oh$^{1,2,10}$, Jinseok Lee$^{2,4}$, Jiyul Lee$^{1,2,3}$, Yechan Seo$^{1,2,3}$,\\
Seong Jeong$^{1,2,3}$, Minsung Kim$^{1,2,4}$, Seonho Shim$^{2,11}$, Younghoon Noh$^{2,4}$, Hyuk Choi$^{1,2,3}$,\\
Youngbin Kong$^{1,9}$, Kyu Eun Lee$^{\dagger,6,7}$ and Hyoun-Joong Kong$^{\dagger,1,3,8}$
\thanks{$^{*}$First author. $^{\dagger}$Corresponding authors.}%
\thanks{$^{1}$Department of Transdisciplinary Medicine, Seoul National University Hospital. $^{2}$Rosota Inc., Seoul, Republic of Korea. $^{3}$Department of Medicine, Seoul National University College of Medicine. $^{4}$Department of Mechanical Engineering, Seoul National University. $^{5}$Department of Computer Science and Engineering, Seoul National University. $^{6}$Department of Surgery, Seoul National University Hospital. $^{7}$Department of Surgery, Seoul National University College of Medicine. $^{8}$Institute of Convergence Medicine with Innovative Technology, Seoul National University Hospital. $^{9}$Interdisciplinary Program in Medical Informatics, Seoul National University College of Medicine. $^{10}$Eulji University College of Medicine. $^{11}$Department of Mechanical Engineering, Chungang University.}%
}

\begin{document}
\AddToShipoutPictureBG*{%
  \put(\LenToUnit{54pt},\LenToUnit{24pt}){%
    \parbox[b]{\textwidth}{\scriptsize This work has been submitted to the IEEE for possible publication. Copyright may be transferred without notice, after which this version may no longer be accessible.}%
  }%
}
\IEEEaftertitletext{\vspace{-0.8in}}   
\maketitle
\thispagestyle{empty}
\pagestyle{empty}

\begin{strip}
\centering
\includegraphics[width=0.960\textwidth]{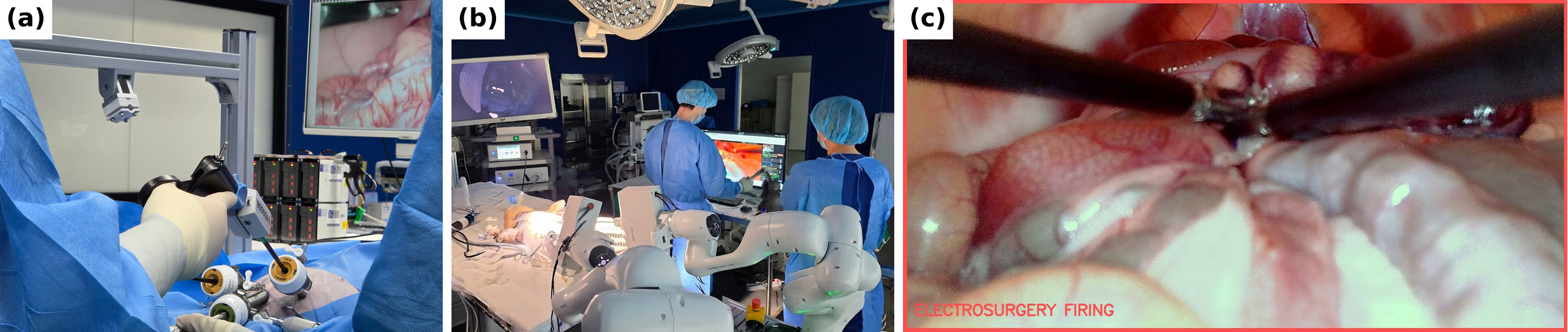}
\captionof{figure}{\textbf{Overview: from hand-held demonstrations to in-vivo robotic execution.} (a)~A surgeon performs laparoscopic appendectomy on a live rabbit with ordinary hand-held instruments. A shaft-mounted surgical instrument-state logger records the instrument state. (b)~The diffusion policy trained on those recordings performs the same procedure on a different live rabbit with two FR3 arms under the surgeon's supervision. (c)~Endoscope view during policy control at the moment of electrosurgery. Three of four appendectomies were completed under shared autonomy.}
\label{fig:hero}
\end{strip}

\begin{abstract}
Most minimally invasive surgery is still performed with hand-held laparoscopic instruments, and the surgeon's instrument kinematics are lost when the operation ends; only the endoscope video is kept. This paper presents an end-to-end pipeline that captures this motion in the operating room and uses it to train a surgical robot policy, validated on live animals. We introduce a surgical instrument-state logger that mounts on the shaft of a standard laparoscopic instrument and recovers its pose and jaw state from an inertial sensor, a time-of-flight sensor and a Hall sensor, with no external camera or tracker. A data pipeline measures the latency of every sensor channel against a robot ground truth and aligns the channels before forming observation--action pairs. On these demonstrations we train a diffusion policy with a fine-tuned DINOv3 backbone, selecting its design by closed-loop rollouts in a physics simulator reconstructed from depth maps of an ex-vivo rabbit appendix. The policy is then retrained on 849 in-vivo demonstrations from four live rabbits and deployed on four additional live rabbits with electrosurgery armed. With the surgeon selecting the surgical phase, the policy completed the appendectomy in three of the four animals. The results show that demonstrations recorded from a surgeon's own instruments are sufficient to train, select and deploy a bimanual surgical policy in vivo. The robot serves only as the timing reference for sensor calibration and as the executor, and collects no demonstrations. Both demonstration corpora are released to support future surgical robot learning research.
\end{abstract}

\section{INTRODUCTION}

Most minimally invasive surgery is still performed by hand. Robotic surgery is growing, with about 3.15 million da~Vinci procedures worldwide in 2025~\cite{intuitive2026results}, yet in a US national inpatient sample of 2012--2019 admissions three of four minimally invasive abdominal operations were laparoscopic rather than robotic (75.8\% vs.\ 24.2\%)~\cite{ng2023national}. On a teleoperated robot such as the da~Vinci, the surgeon's instrument motion is recorded as a by-product; in hand-held surgery, everything except the endoscope video is lost when the operation ends. That lost signal is exactly what imitation learning needs.

Imitation learning has made bimanual manipulation policies practical in everyday manipulation settings~\cite{chi2023diffusion,zhao2023act}, and surgical task autonomy~\cite{attanasio2021autonomy} has been shown in vivo with engineered pipelines~\cite{saeidi2022star}, ex vivo with policies learned from teleoperation~\cite{kim2025srth}, and in vivo with simulator-trained policies for assistive subtasks~\cite{long2025embodied}. Surgery has lagged because its demonstrations had to be collected on the robot itself~\cite{gao2014jigsaws,kim2024srt}, in small corpora tied to one platform. This paper asks whether the discarded signal of hand-held laparoscopy can be captured in the operating room and turned into a surgical robot policy, and answers it end to end for laparoscopic appendectomy. Our thesis is that a small instrument-state logger on the surgeon's ordinary instrument can stand in for the robot at demonstration time if its channels are synchronized to within one policy step, and that every later stage, from data processing to in-vivo retraining, can be carried out on demonstrations the robot never collected. Our contributions span the entire pipeline:

\begin{enumerate}
    \item \textbf{Surgical instrument-state logger} (Sec.~\ref{sec:hardware}--\ref{sec:pairs}), an operating-room-ready device mounted on the instrument that recovers the full instrument state from an inertial measurement unit (IMU), a time-of-flight sensor and a Hall sensor under the trocar's remote-center-of-motion constraint, and a processing chain built on \emph{per-channel} latency matching against a robot ground truth. The matching halves the dynamic depth error and, when ablated in closed loop, proves necessary for the lift stage (81\% vs.\ 19\%).
    \item \textbf{A policy configuration selected by closed-loop evidence} (Sec.~\ref{sec:protocol}--\ref{sec:exvivo}): instead of choosing the policy design by offline validation error, we compare 31 candidate designs (backbone, unfreezing depth, phase input, latency matching, replanning) by closed-loop rollouts in a physics simulator built from depth recordings of the ex-vivo scene, score them by stage counts and exact tests, and run only the winner on a two-FR3 rig. A frozen self-supervised ViT backbone removes an order of magnitude of unsafe behavior relative to the standard ResNet-18 encoder, and offline action error turns out to be anti-correlated with closed-loop safety.
    \item \textbf{In-vivo execution on live animals} (Sec.~\ref{sec:invivo}): the configuration retrained from scratch on 849 in-vivo demonstrations completed the appendectomy in three of four live rabbits under shared autonomy with electrosurgery armed. Ex-vivo pre-training did not change a single in-vivo decision, which is itself the argument for a device that collects in vivo.
\end{enumerate}

We do not claim full autonomy~\cite{yang2017medical}: the surgeon selected the surgical phase because the vision-based phase predictor collapsed under the deployment endoscope (Sec.~\ref{sec:discussion}). Nor a robot-free pipeline: an FR3 is the ground truth for the latency measurement (Sec.~\ref{sec:latency}) and the ex-vivo test rig, but collected no training demonstrations.

\section{RELATED WORK}
\label{sec:related}

\textbf{Learning surgical manipulation.} The STAR program reached in-vivo task autonomy for anastomosis with engineered planners and markers~\cite{saeidi2022star}, and autonomous electrosurgery has been shown with marker-based tissue tracking~\cite{saeidi2019electrosurgery}. Learned surgical policies have been trained from teleoperation on the da~Vinci Research Kit~\cite{kazanzides2014dvrk}, from corpora such as JIGSAWS~\cite{gao2014jigsaws}, and up to the imitation of procedural steps ex vivo~\cite{kim2024srt,kim2025srth}. Long et al.~\cite{long2025embodied} brought policies trained by reinforcement learning in simulation to a live animal for three assistive tasks (retraction, gauze picking, clipping), without energy. We differ in demonstration source (a surgeon's hand-held instruments, not teleoperation or a simulator), task (the steps of an operation, ending in tissue division with energy) and platform (general-purpose arms with standard instruments).

\textbf{Robot-free demonstration collection.} Hand-held capture devices such as Dobb-E's stick~\cite{shafiullah2023dobbe} and UMI~\cite{chi2024umi} supply policy training data without a robot. UMI showed that a hand-held gripper can train a diffusion policy if latency is measured and compensated; we extend its compensation to \emph{heterogeneous} channels whose delays differ by an order of magnitude. Trackers on hand-held laparoscopic instruments have measured motion for skill assessment since the 2000s~\cite{chmarra2007tracking}, but never as synchronized, jaw-inclusive policy demonstrations. For those, the robot-free source explored so far is the video itself: SurgiPose~\cite{chen2025surgipose} and the SurgVIL preprint~\cite{chen2026surgvil} estimate instrument kinematics from monocular video by fitting a CAD model under a stationary endoscope and train dVRK policies on them, at 6--12\,mm position error, and even the best markerless estimators on the SurgRIPE and SurgPose benchmarks~\cite{xu2025surgripe,wu2025surgpose} reach 2.6--3.0\,mm, and only without occlusion; with occlusion the error is 5.4--12.4\,mm~\cite{xu2025surgripe}. On our own rig, against the FR3 tip pose (Sec.~\ref{sec:latency}), a markerless fit of the shaft's two silhouette edges in the undistorted endoscope image located the shaft axis only to a median 10.7\,mm, whereas the ToF and IMU of our instrument-mounted logger, scored against the same FR3 truth during free motion, gave 1.36\,mm in depth and 0.17--0.42$^\circ$ in attitude. Since a 2.7\,mm depth error already removes the lift stage (Sec.~\ref{sec:exvivo}), we put a sensor on the instrument instead.

\begin{figure}[t]
\centering
\includegraphics[width=0.9\columnwidth]{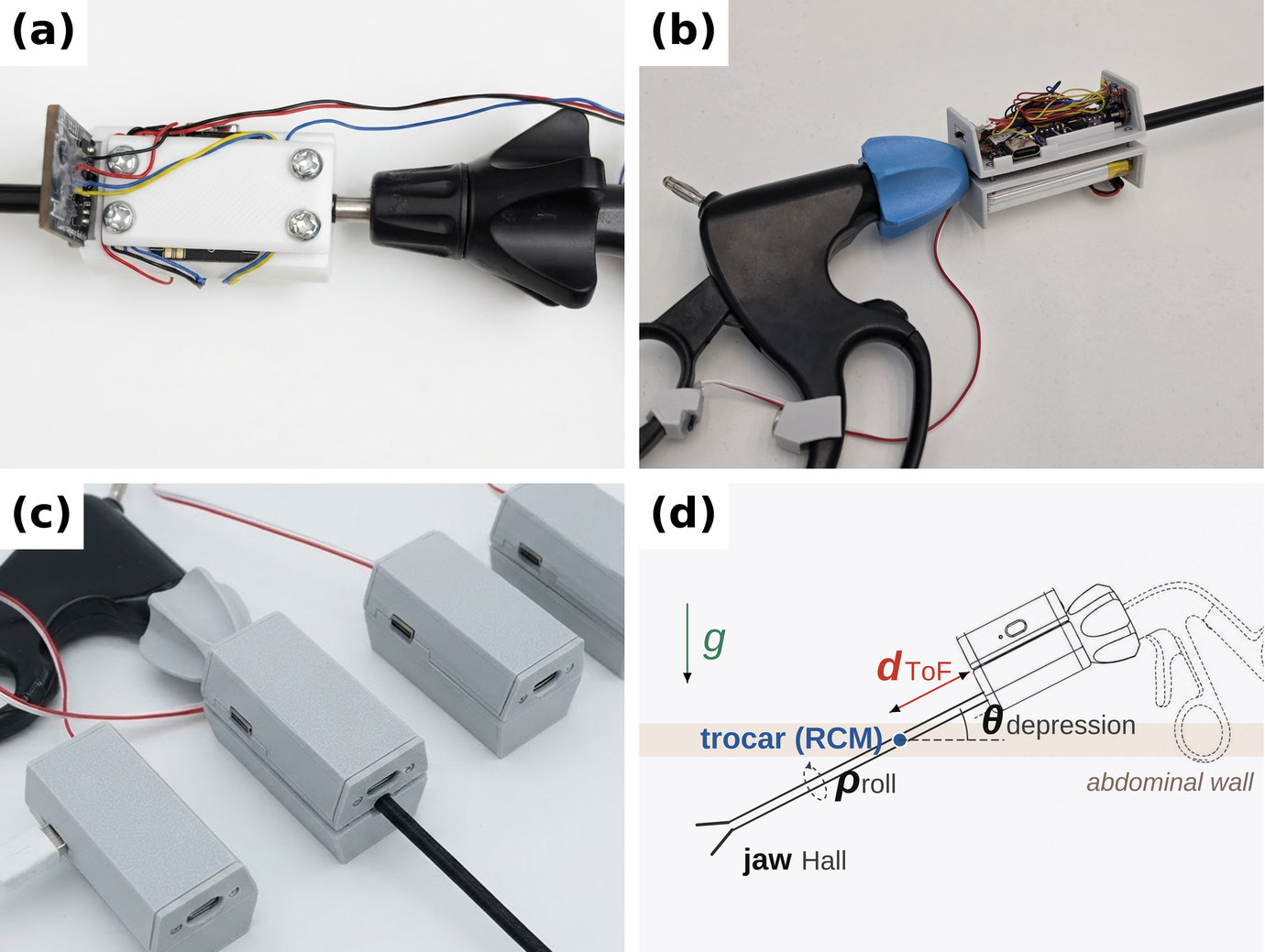}
\caption{\textbf{The surgical instrument-state logger.} (a)~The ex-vivo logger on a laparoscopic grasper. (b)~The in-vivo logger with its enclosure open: nRF52840 microcontroller, BNO085 IMU, VL53L1X ToF sensor, 1000\,mAh lithium-polymer cell and, on the handle, the WSH137 Hall sensor and its magnet. (c)~Four in-vivo loggers as used in the operating room: the same sensors in a compact enclosure with no exposed wiring. (d)~Geometry: four degrees of freedom through the trocar. Roll~$\rho$ and depression~$\theta$ come from the gravity-referenced IMU, insertion depth~$d$ from the ToF ray to the trocar, and jaw opening from the Hall sensor.}
\label{fig:hardware}
\end{figure}

\textbf{Visuomotor policies and backbones.} We build on Diffusion Policy~\cite{chi2023diffusion} (DDPM~\cite{ho2020ddpm}, DDIM~\cite{song2021ddim}, action chunking~\cite{zhao2023act}, FiLM~\cite{perez2018film}) with DINOv3~\cite{simeoni2025dinov3}, the successor of DINOv2~\cite{oquab2024dinov2}, in place of its ResNet-18 encoders. A surgical backbone (SurgeNetXL~\cite{jaspers2025surgenetxl}) gave no closed-loop benefit. Among vision--language--action models~\cite{zitkovich2023rt2,black2024pi0} we fine-tune GR00T-H, the medical-robotics VLA released with the Open-H-Embodiment dataset~\cite{nelson2026openh}, as a baseline. Phase recognition from video follows TeCNO~\cite{czempiel2020tecno}.

\textbf{Surgical simulators and sim-to-real.} Existing learning simulators target the dVRK~\cite{xu2021surrol,scheikl2023lapgym,yu2024orbit}, whereas ours is an Isaac Sim reconstruction of the phantom in which the demonstrations were collected. Because sim-to-real transfer for deformable tissue remains unreliable~\cite{scheikl2023simtoreal}, we use it to compare configurations under one protocol and run only the chosen one on the real rig.

\section{METHODS}
\label{sec:methods}

\subsection{The surgical instrument-state logger}
\label{sec:hardware}
\begin{figure}[t]
\centering
\includegraphics[width=0.86\columnwidth]{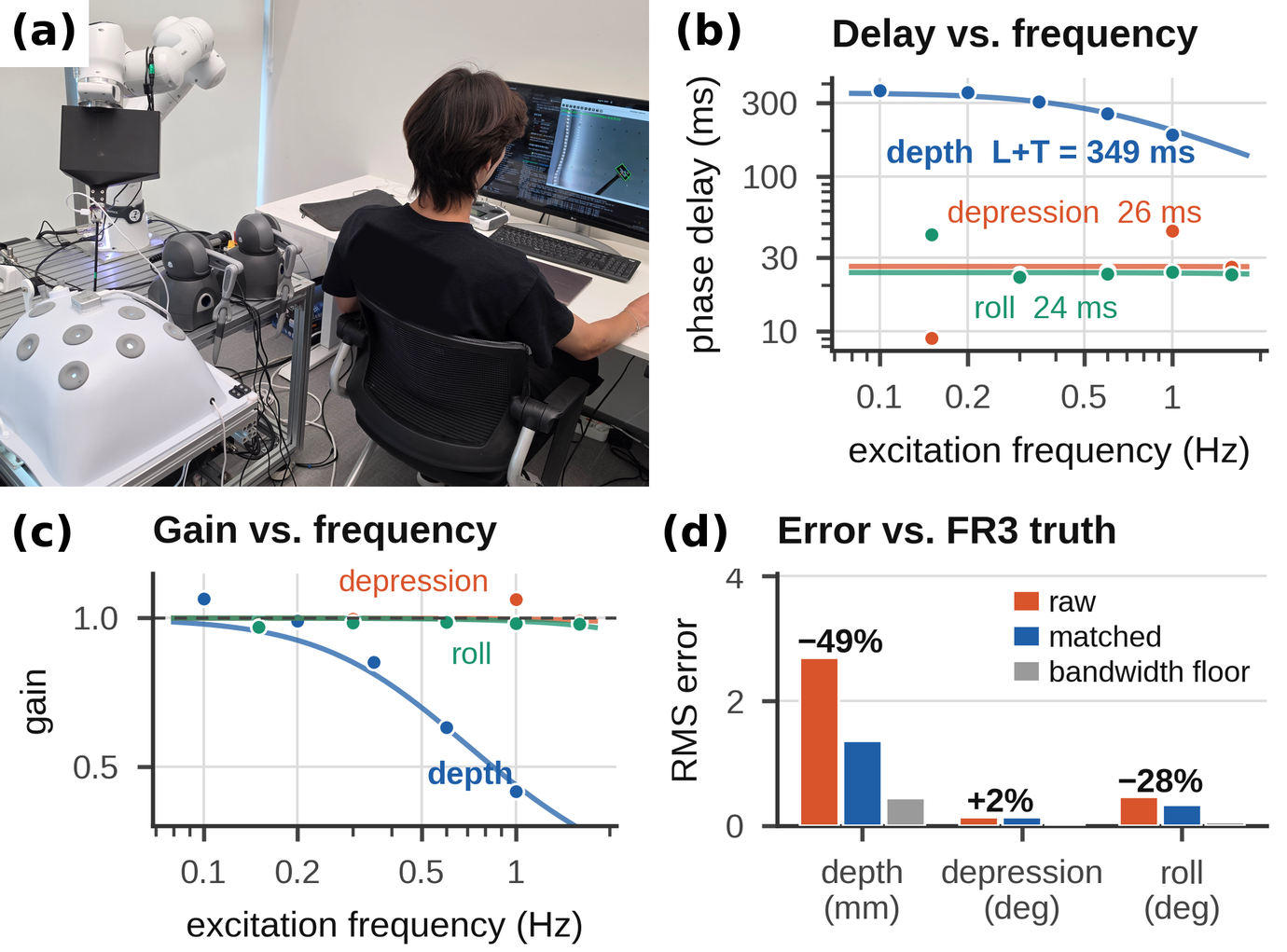}
\caption{\textbf{Per-channel latency.} (a)~An FR3 drives the instrument through the phantom trocar with swept sines; its tip pose is the ground truth. (b)~Fitted delay and (c)~gain per channel versus excitation frequency, ex-vivo logger (in-vivo logger: same protocol, values in the text). Only insertion depth is frequency dependent, a first-order lag rather than a transport delay. (d)~Dynamic error before and after rewinding each channel by its own latency; the depression channel is the control.}
\label{fig:latency}
\end{figure}
\begin{figure}[t]
\centering
\includegraphics[width=0.96\columnwidth]{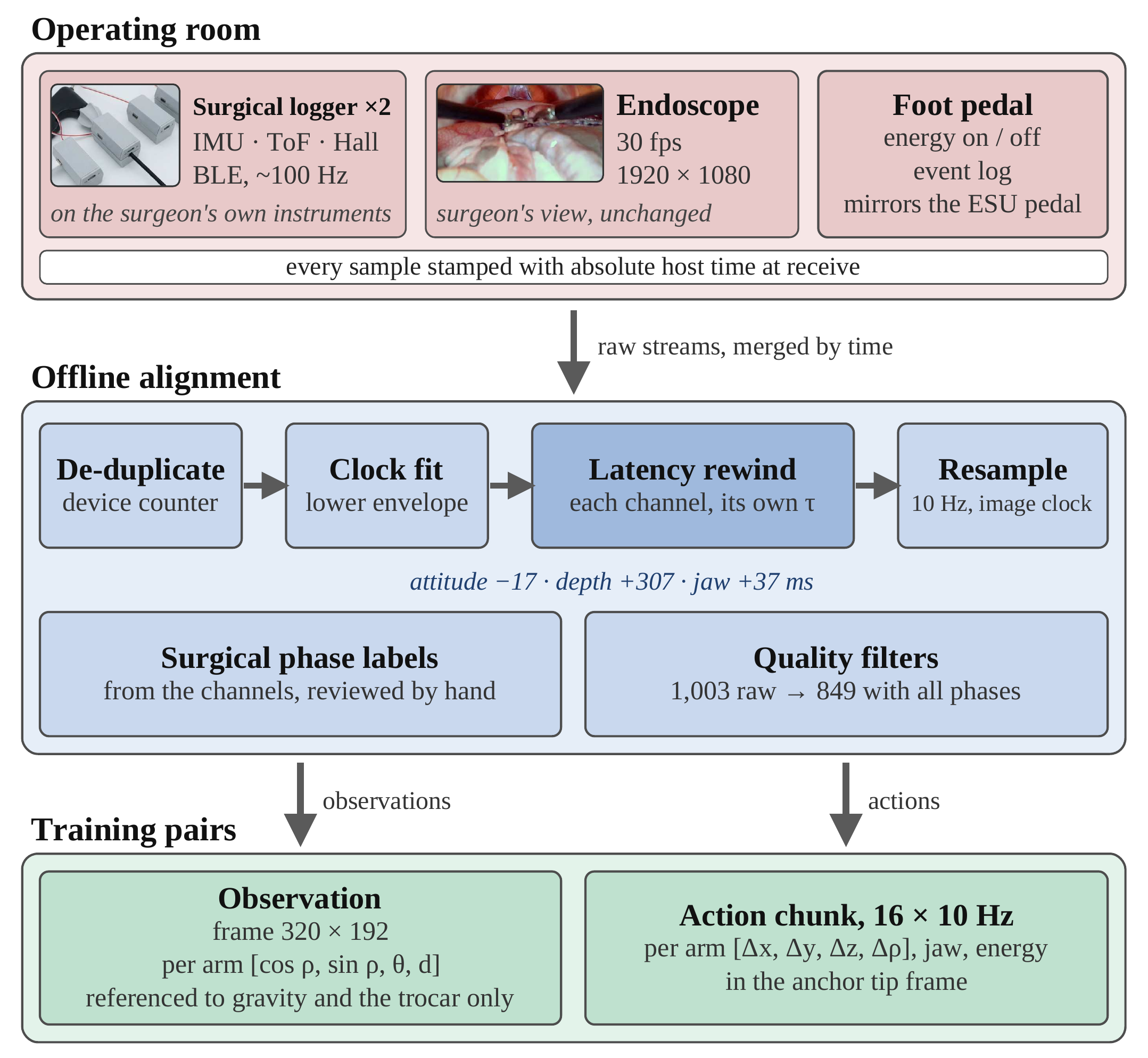}
\caption{\textbf{Data pipeline from raw sensor logs to training pairs.} Every stream is stamped with absolute time, rewound by its own measured latency and resampled on the endoscope clock. Observations use only quantities referenced to gravity and the trocar, so that magnetometer drift never enters the policy.}
\label{fig:pipeline}
\end{figure}
\begin{figure*}[t]
\centering
\includegraphics[width=0.902\textwidth]{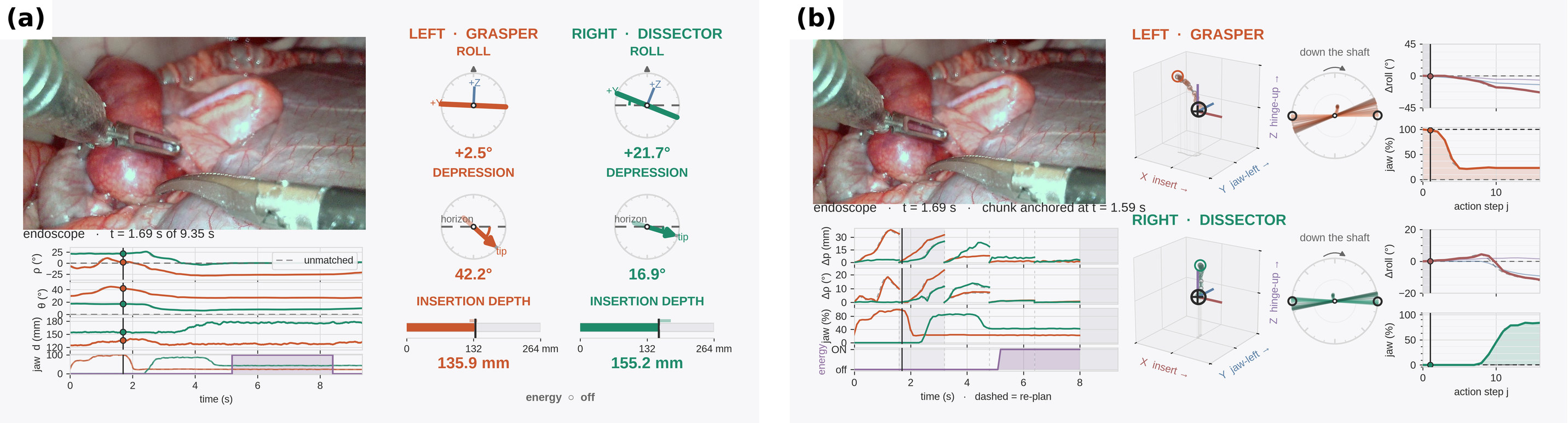}
\caption{\textbf{Policy observation and action chunk} for one in-vivo demonstration at $t{=}1.7$\,s, as the grasper closes on the appendix and the dissector approaches. (a)~Model input: the endoscope frame and, per arm, shaft roll~$\rho$, depression~$\theta$ and insertion depth~$d$ after per-channel latency matching, as time series and as instantaneous gauges. (b)~The action chunk anchored at 1.6\,s: each arm's 16-step tip path in its tool frame, the emitted roll (spokes) and jaw channels, and the energy timeline (mauve). All of it was recorded by the surgical instrument-state logger on the surgeon's own instruments.}
\label{fig:io}
\end{figure*}
The surgical instrument-state logger records the pose and jaw opening of a hand-held laparoscopic instrument during real operations, turning the surgeon's own motion into demonstration data. It mounts on the shaft outside the patient, needs no external camera or tracker, and leaves the part that enters the patient untouched. Against the FR3 tip pose during free motion (Sec.~\ref{sec:latency}), its latency-matched channels track shaft roll and depression to 0.42$^\circ$ and 0.17$^\circ$ RMS and insertion depth to 1.36\,mm RMS once the firmware's fixed gain and offset are calibrated out (5.6\,mm before), and the tip displacement over a 1.6\,s action chunk to 1.38\,mm RMS (2.89\,mm at the 95th percentile), whereas video-based instrument tracking reaches several millimeters (Sec.~\ref{sec:related}). A rigid instrument through a trocar has four degrees of freedom, not six, because the remote-center-of-motion (RCM) constraint removes two (Fig.~\ref{fig:hardware}(d)). Orientation about the trocar comes from an IMU. The remaining degree of freedom, insertion depth, is $d=\ell-r$: the shaft length $\ell$ from the board to the tip minus the board-to-trocar distance $r$, which a time-of-flight (ToF) sensor pointed down the shaft measures directly. Each instrument therefore carries one board (Seeed XIAO nRF52840) with a BNO085 IMU and a ToF sensor, plus a linear Hall sensor on the handle (WSH137-XPAN1, 4.4\,mV/G, sampled at 200\,Hz) that reads a magnet on the moving handle ring and so measures jaw opening (Fig.~\ref{fig:hardware}).

Two generations share this principle. The ex-vivo logger (VL53L0X ToF) was powered and read over a USB-C cable to the recording host and smoothed the depth with an exponential moving average ($\alpha{=}0.1$). The in-vivo logger was redesigned for the operating room: a faster ToF (VL53L1X, five-sample moving average, 40\,ms), a 1000\,mAh lithium-polymer cell and BLE streaming at 50--200\,Hz, so that no cable leaves the instrument, all housed in a compact enclosure (Fig.~\ref{fig:hardware}(b),(c)). At the measured draw of about 50\,mA the cell lasts roughly 20\,h, beyond one operating-room day.

\textbf{Synchronization.} Every recorder---the boards, the endoscope capture and a foot-pedal ``energy'' channel that mirrors the electrosurgical pedal---stamps each sample with the host's absolute time at the earliest receive point; streams are merged offline by that time. The board clock is mapped to host time by a \emph{lower-envelope} regression because arrival jitter is one-sided.

\subsection{Per-channel latency matching}
\label{sec:latency}
Each channel reaches the log later than the event it describes, by a different amount. A Franka FR3 drove an instrument carrying the surgical instrument-state logger through the phantom trocar along swept-sine trajectories while its tip pose, published at 200\,Hz from the 1\,kHz joint state, served as ground truth (Fig.~\ref{fig:latency}(a)), and for each channel we fitted the response
\begin{equation}
H(j\omega)=e^{-j\omega L}/(1+j\omega T),
\end{equation}
where $L$ is a pure transport delay and $T$ the time constant of a first-order lag, whose delay grows and gain falls with frequency. The low-frequency delay is $L{+}T$. A channel with $T{\approx}0$ can be corrected by a time shift; one with a large $T$ cannot, because part of the motion is lost.

On the ex-vivo logger the attitude channels showed pure transport delays of 24--26\,ms. Insertion depth showed $L{+}T{=}349$\,ms, of which 93\% was first-order lag from the firmware's exponential moving average, with gain falling to 0.42 at 1\,Hz (Fig.~\ref{fig:latency}(b),(c)). The jaw channel's 79\,ms came from the firmware loop and the endoscope's 42\,ms from cross-correlation with IMU angular speed. The 325\,ms depth--attitude skew is 3.5 policy steps at 10\,Hz; no common time shift can remove it. Following UMI's sample-ahead rule we rewind each channel by its own latency and interpolate onto the image clock ($-17$, $+307$ and $+37$\,ms for attitude, depth and jaw). Against the FR3 truth this reduced dynamic depth error from 2.69 to 1.36\,mm ($-49.5\%$) and shaft-roll error by 28\%. The depression channel, whose 26\,ms latency is below what the motion resolves, changed by only $+2\%$ and serves as a control (Fig.~\ref{fig:latency}(d)). The in-vivo logger, measured with the same FR3 protocol, showed depth latencies of 57 and 63\,ms (two batches), attitude 3--9\,ms and a computed jaw 43\,ms, so the endoscope's own 55\,ms (cross-correlation pooled over 196 in-vivo episodes) became the dominant term.

\subsection{From logs to training pairs}
\label{sec:pairs}
The magnetometer heading of a low-cost IMU drifts in an operating room, so the observation uses only what gravity and the trocar define (Figs.~\ref{fig:pipeline} and~\ref{fig:io}(a)): the endoscope frame ($320\times192$) and, per arm, $[\cos\rho,\sin\rho,\theta,d/d_{\max}]$---shaft roll and depression from the gravity-referenced attitude, and insertion depth from the ToF. The jaw aperture is deliberately \emph{not} an input: with it the policy learns ``it is closed now, so keep closing''. Actions are 16-step chunks at 10\,Hz (Fig.~\ref{fig:io}(b)) of per-arm tip-frame deltas $(\Delta x,\Delta y,\Delta z,\Delta\rho)$ relative to the chunk's anchor pose, absolute jaw aperture, and, in vivo, an absolute electrosurgery channel owned by the right (cutting) arm. The deltas come from the same channels: at each step the full IMU attitude and the ToF depth place the tip on the shaft line through the trocar, and the tip poses of a chunk are then expressed in the tool frame of its first step. Heading drift is negligible within one 1.6\,s chunk (0.02--0.06$^\circ$ of systematic error, under 5\% of the roll range), so the relative motion is trusted where the absolute heading is not. Because every label is relative to its anchor pose, it is invariant to the heading zero and to the trocar location, and at deployment the executor unwinds the chunk in the robot's own measured tip frame, so no correspondence between the demonstration trocar and the FR3 trocar is needed.

\subsection{Policy}
\begin{figure*}[t]
\centering
\includegraphics[width=0.960\textwidth]{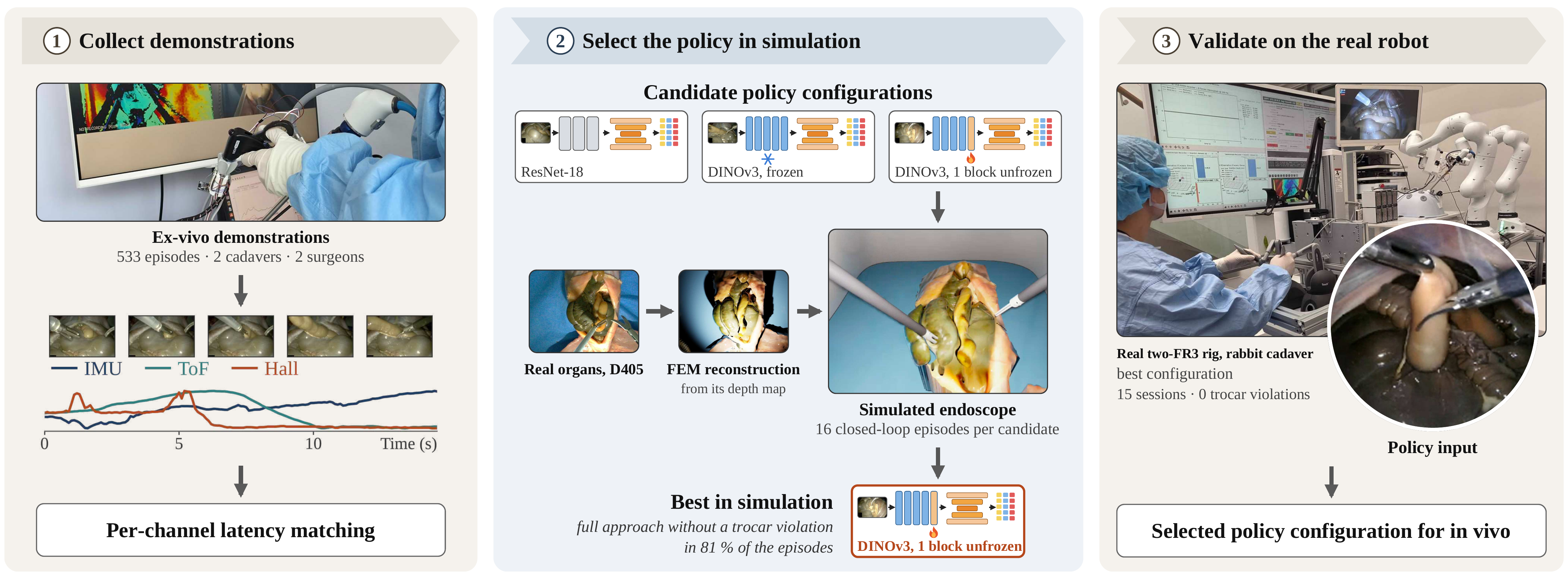}
\caption{\textbf{Ex-vivo policy selection.} Demonstrations recorded with the ex-vivo logger in a laparoscopic phantom (left) train the candidate policy configurations. Every candidate runs closed-loop episodes in a simulator whose organs are reconstructed from the phantom's depth maps (middle). The configuration with the highest full-approach rate without a trocar violation is run on the real two-FR3 rig (right) and becomes the policy configuration retrained for in vivo.}
\label{fig:venues}
\end{figure*}
\label{sec:policy}
We use a 1-D U-Net diffusion policy~\cite{chi2023diffusion,ho2020ddpm} with DDIM sampling~\cite{song2021ddim} (8 steps) over 16-step action chunks~\cite{zhao2023act}, $n_{\mathrm{obs}}{=}2$ (0.2\,s), and three changes selected by the experiments of Sec.~\ref{sec:exvivo}. We call such a set of design choices a \emph{policy configuration}.

\textbf{Backbone.} A single shared DINOv3 ViT-B/16~\cite{simeoni2025dinov3} replaces the two ResNet-18 encoders, with attention pooling and only the last transformer block trainable (7.1\,M of 85.7\,M parameters).

\textbf{Phase conditioning.} Because the observation window is 0.2\,s, the policy cannot infer where in the procedure it is, so we condition the U-Net by FiLM~\cite{perez2018film} on an ordinal encoding of the surgical phase (grasp approach, lift, cut approach, cut) plus two timing channels, the fraction of the phase elapsed and the normalized time to the next boundary. At training time all three come from labels derived from the instrument channels and corrected by hand for every episode. At run time they come from a vision-based predictor: a causal temporal convolutional network (TCN) that reads features extracted by the frozen backbone from recent endoscope frames and regresses the phase and both timing channels. Because the phases occur in a fixed order, its phase output may only advance. When the surgeon supplies the phase by keypress, only the phase is replaced. The timing channels remain the predictor's video estimates, so nothing unavailable at run time enters the policy. In simulation the \emph{manual phase} and both timing channels are read from the simulator state.

\textbf{Executor and latency budget.} A chunk executor commits $n_{\mathrm{exec}}$ actions from index $j_0$ of each chunk and replans; following UMI's latency-budget rule~\cite{chi2024umi}, $j_0$ absorbs the observation anchor, the arm's execution latency and inference. The arm's response was measured on the FR3 with swept sines as a pure delay $L{=}85$\,ms plus a first-order time constant $T{=}205$\,ms, and only $L$ is advanced. In vivo this gives $j_0{=}\{2,2,1,1\}$ for tip, roll, jaw and energy with $n_{\mathrm{exec}}{=}3$ at 5\,Hz, because the in-vivo demonstrations move at up to 110\,mm/s (p99) against a 50\,mm/s controller limit. Electrosurgery activation must pass five gates: an activation threshold, two consecutive requests, a closed jaw on the cutting instrument, an armed state, and an operator interlock.

\subsection{Simulator and selection protocol}
\label{sec:protocol}
Every candidate configuration is evaluated in closed loop in an Isaac Sim reconstruction of the phantom: six FEM organs meshed from the ex-vivo RealSense~D405 recording, nominal soft-tissue stiffness (Young's modulus 12--30\,kPa, Poisson's ratio 0.45, not fitted to the specimen), cameras at the poses measured on the real rig, and two FR3s through 6\,mm trocars (Fig.~\ref{fig:venues}). Every policy is scored under one fixed protocol: 16 episodes of 15\,s, evaluated on stage counts (grasp, lift $\geq$10\,mm, at site, jaw on tissue, cut) and on one hard failure, a \emph{trocar violation}. The simulated RCM error, the distance between the shaft and the trocar point, grows when the policy pushes hard on tissue or drives the instrument into the phantom or the floor. A maximum above 10\,mm ends the episode as a violation. With 16 episodes per configuration we report stage counts with Wilson intervals and compare configurations with Fisher's exact test.

\section{EXPERIMENTS}

\subsection{Demonstration corpora}
\label{sec:corpora}
Table~\ref{tab:data} lists the two corpora. Both were captured with the surgical instrument-state logger alone: one unit on the shaft of each instrument streamed its channels to the recording host (over USB ex vivo, over BLE in vivo), which also captured the endoscope and the pedal channel and stamped every sample on arrival (Sec.~\ref{sec:hardware}). No marker, external tracker or calibration object entered the field, and the surgeon worked with the instruments as usual. Each episode is one attempt at the procedure.

\textbf{Ex-vivo.} Rabbit cadavers were used on three session days: each was opened, the appendix exposed and placed inside a laparoscopic training phantom, and the appendectomy performed repeatedly with the ex-vivo logger (Fig.~\ref{fig:venues}). The cut was rehearsed without dividing tissue, which was then divided once at the end. A RealSense~D405 above the phantom recorded the field, the source of the simulator's organ geometry. The training corpus is the second day alone: 533 episodes (74\,min) on two specimens, recorded by two surgeons in a morning and an afternoon session.

\textbf{In-vivo.} Eight live rabbits were used in total, four for demonstrations and four for deployment. All animal procedures involving rabbits were approved by the BIOSTEP Institutional Animal Care and Use Committee (IACUC S26-BZ-0562). In one operating-room session, the four demonstration animals received trocars directly through the abdominal wall, with no phantom (Fig.~\ref{fig:invivo}(a)). The cutter was replaced by a dissector and an electrosurgical unit was connected. Each animal's cut was rehearsed in several episodes with the pedal but without current, and one final episode divided the appendix with energy. Of 1,003 raw episodes, 849 survive the filters of Sec.~\ref{sec:invivotrain}. Both training corpora are released for reproducibility at \url{https://github.com/Rosota-Research/ICRA2027-Invivo-main}.

\begin{table}[t]
\centering
\caption{\textbf{Demonstration corpora used for training.} Each episode is one attempt at the appendectomy recorded with the surgical instrument-state logger on the surgeon's hand-held instruments. Duration is the total recorded time after quality filtering.}
\label{tab:data}
\small
\setlength{\tabcolsep}{4pt}
\begin{tabular}{@{}lrr>{\raggedright\arraybackslash}p{3.6cm}@{}}
\toprule
Corpus & Episodes & Duration & Setting \\
\midrule
Ex-vivo & 533 & 74\,min & 2 rabbit cadavers in a phantom, ex-vivo logger \\
In-vivo & 849 & 135\,min & 4 live rabbits, in-vivo logger, energy pedal \\
\bottomrule
\end{tabular}
\end{table}

\subsection{Configuration selection on the ex-vivo corpus}
\label{sec:exvivo}
Table~\ref{tab:ablation} summarizes the design axes that mattered.

\begin{table*}[t]
\centering
\caption{\textbf{Closed-loop simulator results of the ex-vivo configuration ablation.} 16 episodes of 15\,s per configuration. Lift and Full approach (the selection criterion) count violation-free episodes. Italic rows name the factor varied and the settings already fixed. Shaded rows mark the setting kept. All rows use latency matching and $n_{\mathrm{exec}}{=}3$ unless varied.}
\label{tab:ablation}
\footnotesize\setlength{\tabcolsep}{3pt}\renewcommand{\arraystretch}{0.95}\setlength{\aboverulesep}{1.54pt}\setlength{\belowrulesep}{0.46pt}
\dimen0=\ht\strutbox \advance\dimen0 by 0.57pt
\dimen2=\dp\strutbox \advance\dimen2 by -0.57pt
\setbox\strutbox=\hbox{\vrule height\dimen0 depth\dimen2 width 0pt}
\begin{tabularx}{\dimexpr\textwidth-2\tabcolsep\relax}{@{}>{\raggedright\arraybackslash}X l c l r r@{}}
\toprule
 & & \multicolumn{2}{c}{Progress, violation-free episodes} & Grasp & \\
\cmidrule(lr){3-4}
Configs & Trocar violation $\downarrow$ [95\% CI] & Lift $\uparrow$ & Full approach $\uparrow$ [95\% CI] & depth (mm) $\uparrow$ & MAE $\downarrow$ \\
\midrule
\multicolumn{6}{@{}l}{\textit{Model}} \\
GR00T-H (VLA), no phase & 12\% [3--36] & 50\% & 0\% [0--19] & --- & 0.184 \\
DP, ResNet-18 & 88\% [64--97] & 12\% & 12\% [3--36] & 13.9 & 0.145 \\
DP, DINOv3 frozen & \textbf{6\% [1--28]} & 69\% & 44\% [23--67] & 0.0 & 0.152 \\
\rowcolor{keptblue}DP, DINOv3, last block unfrozen & 19\% [7--43] & \textbf{81\%} & \textbf{81\% [57--93]} & 10.8 & 0.148 \\
\midrule
\multicolumn{6}{@{}l}{\textit{Phase input}\ \ (fixed: DINOv3, last block unfrozen)} \\
\rowcolor{keptblue}Manual phase & 19\% [7--43] & \textbf{81\%} & \textbf{81\% [57--93]} & 10.8 & 0.148 \\
TCN (demo video) & 50\% [28--72] & 31\% & 25\% [10--49] & 5.2 & 0.148 \\
TCN (simulator) & 25\% [10--49] & 38\% & 38\% [18--61] & 10.3 & 0.148 \\
\midrule
\multicolumn{6}{@{}l}{\textit{Unfrozen DINOv3 blocks}\ \ (fixed: manual phase)} \\
None (frozen) & \textbf{6\% [1--28]} & 69\% & 44\% [23--67] & 0.0 & 0.152 \\
\rowcolor{keptblue}Last 1 block & 19\% [7--43] & \textbf{81\%} & \textbf{81\% [57--93]} & 10.8 & 0.148 \\
Last 2 blocks & 19\% [7--43] & 50\% & 50\% [28--72] & 10.7 & 0.146 \\
Last 3 blocks & 56\% [33--77] & 38\% & 38\% [18--61] & 9.3 & 0.143 \\
Last 4 blocks & 69\% [44--86] & 31\% & 31\% [14--56] & 8.8 & 0.142 \\
\midrule
\multicolumn{6}{@{}l}{\textit{Latency matching}\ \ (fixed: DINOv3, last block unfrozen, manual phase)} \\
\rowcolor{keptblue}With & 19\% [7--43] & \textbf{81\%} & \textbf{81\% [57--93]} & 10.8 & 0.148 \\
Without & 19\% [7--43] & 19\% & 19\% [7--43] & 11.0 & 0.138 \\
\midrule
\multicolumn{6}{@{}l}{\textit{Replanning}\ \ (fixed: DINOv3, last block unfrozen, manual phase)} \\
\rowcolor{keptblue}$n_{\mathrm{exec}}{=}3$ & 19\% [7--43] & \textbf{81\%} & \textbf{81\% [57--93]} & 10.8 & 0.148 \\
$n_{\mathrm{exec}}{=}1$ & 69\% [44--86] & 31\% & 31\% [14--56] & 10.7 & 0.148 \\
\bottomrule
\end{tabularx}

\\[2pt]
\raggedright\footnotesize DP: diffusion policy.
\end{table*}

\textbf{Selection criterion.} A policy that does nothing never violates the trocar, so violations alone cannot select a configuration; GR00T-H and the frozen backbone score well on it for that reason. The criterion was therefore the full approach without a violation---lift, arrive at the site and close the jaw on tissue---the furthest stage the simulator can score, reached with few violations. On this criterion the DINOv3 configuration with its last transformer block unfrozen was the best of the 31 (81\%), ahead of two unfrozen blocks (50\%) and of the fully frozen backbone (44\%), whose low violation rate comes at the price of never closing the jaw on tissue (grasp depth 0.0\,mm). GR00T-H~\cite{nelson2026openh} (0\%, none of 24 checkpoints closed a jaw on tissue) and ResNet-18 ($\le$12\%) are far behind. The cost is reduced trocar safety in the simulator (19\% vs.\ 6\% violations, $p=0.6$), a cost that did not materialize on the real rig or in vivo.

\textbf{Backbone.} Replacing the two ResNet-18 encoders with a frozen DINOv3 reduced trocar violations from 88\% to 6\% (Fisher $p=7\times10^{-6}$) and raised lift from 12\% to 69\% ($p=0.003$).

\textbf{Latency matching.} Training the deployed configuration on the same episodes processed \emph{without} per-channel latency matching leaves trocar violations unchanged (19\% vs.\ 19\%) but eliminates successful lifts (19\% vs.\ 81\%, $p=0.001$), while its validation mean absolute action error (MAE) is \emph{better} (0.138 vs.\ 0.148): the 325\,ms depth--attitude skew breaks the coordinated insertion that lifting requires, which offline error cannot see.

\textbf{Unfreezing.} Safety degrades as more transformer blocks are unfrozen (violations 6\%, 19\%, 19\%, 56\%, 69\% for zero to four blocks, $p=6\times10^{-4}$ for four vs.\ frozen), while only unfreezing a single block raises the full-approach rate notably above the frozen backbone (81\% vs.\ 44\%, $p=0.07$).

\textbf{Offline error vs.\ closed-loop safety.} That offline validation error does not predict closed-loop success is known from robomimic~\cite{mandlekar2021robomimic}; in our surgical setting it goes further and is \emph{anti}-correlated with safety. Across the unfreezing sweep, validation MAE improves monotonically (0.152$\rightarrow$0.142) while violations rise from 6\% to 69\%. The least safe encoder, ResNet-18, even has a better MAE (0.145) than the frozen DINOv3 (0.152). Capacity (ViT-S+ vs.\ ViT-B), surgical-domain pre-training (SurgeNetXL~\cite{jaspers2025surgenetxl}) and attention pooling on the frozen backbone (violations 6\%$\rightarrow$44\%) show the same inversion.

\textbf{Run-time phase input.} A predictor trained on demonstration video cut the deployed configuration's lift from 81\% to 31\% ($p=0.01$) and its full approach from 81\% to 25\% ($p=0.004$), and collapsed the two-block policy (94\% violations). One trained in the simulator's own domain recovered part of this loss but still halved lift (38\% vs.\ 81\%, $p=0.03$). Removing the phase input from the ResNet policy did not cost progress (full approach 6\% vs.\ 12\%). The phase input matters not because it adds information but because an incorrect phase destroys performance. Replanning matters as much: committing one action per chunk instead of three raised the deployed configuration's violations from 19\% to 69\% and cut its full approach from 81\% to 31\% ($p=0.01$).

\textbf{Real ex-vivo rig.} The deployed configuration ran on the two-FR3 rig (Fig.~\ref{fig:venues}) with a rabbit cadaver (15 sessions, $\approx$18\,min of policy control): zero trocar violations, peak RCM error 0.23\,mm. The phase predictor, however, saturated at the final phase within a median of 10.6\,s (rig frames lay outside its training manifold), so in vivo the surgeon set the phase.

\subsection{In-vivo training}
\label{sec:invivotrain}
\begin{figure*}[t]
\centering
\includegraphics[width=0.634\textwidth]{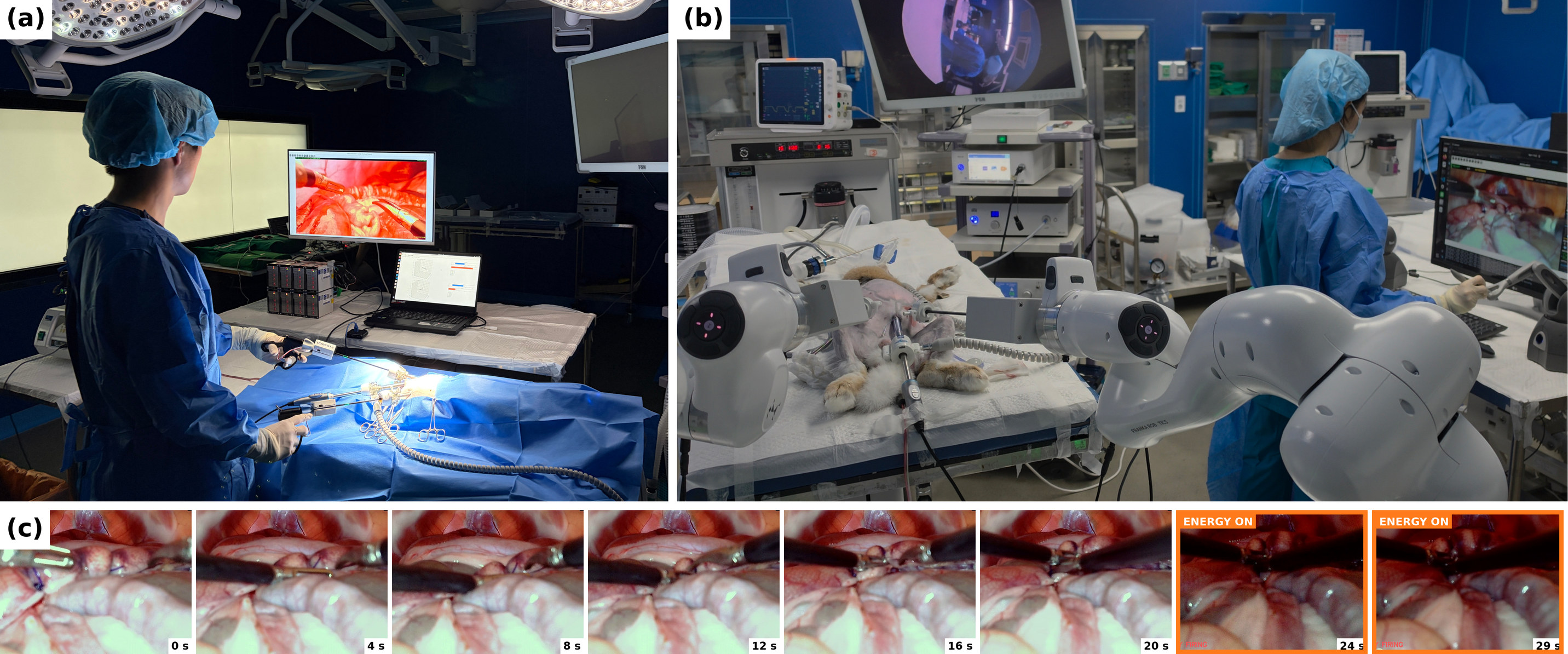}
\caption{\textbf{In-vivo demonstration collection and deployment.} (a)~In-vivo demonstration collection: the surgeon operates on a live rabbit with instruments carrying the in-vivo logger. (b)~Deployment on a different animal: two FR3 arms hold the grasper and dissector through trocars in the anesthetized rabbit with the electrosurgical unit connected. The surgeon supervises at the console. (c)~Frames from rabbit~1's 34\,s policy rollout: grasper approach, lift, dissector approach, jaws closing on tissue, and the energized cut (orange frames, 24--29\,s).}
\label{fig:invivo}
\end{figure*}
The policy was trained \emph{from scratch} on the in-vivo corpus, the 849 demonstrations recorded with the in-vivo surgical instrument-state logger (Sec.~\ref{sec:corpora}). Of 1,003 raw episodes, 135 were dropped for sensor-stream quality (packet loss above 3\%, rate below 90\,Hz, truncated streams) or because they were not part of the procedure (animal preparation, a sensor test), and 19 more for a missing phase or an early-ending stream. The ex-vivo configuration and optimizer-step budget were unchanged. Because offline error had failed as a selector, the checkpoint was accepted through pre-registered event-level gates on the last 5\% of episodes in recording order (unseen episodes): every jaw-closing event detected, energy false-fire 0.019 and missed-fire 0.032 (gates 0.05 and 0.30), jaw timing within one anchor, and closed-loop gain 0.86/0.82 (left/right) against a 0.8 gate.

\textbf{Effect of ex-vivo pre-training.} We trained the in-vivo policy from random initialization and from the deployed ex-vivo weights, on 25\%, 50\% and 100\% of the in-vivo corpus with the same step budget. The two initializations passed the same pre-registered gates in all six runs, and in offline error (in-vivo units) ex-vivo pre-training helped only while data were scarce: MAE 1.221 vs.\ 1.413 at 25\% of the corpus, 1.176 vs.\ 1.173 at 50\%, and 1.157 vs.\ 1.173 at 100\%. A policy for live tissue must be trained on live-tissue demonstrations, which is why an instrument that collects them in the operating room is a necessity, not a convenience.

\subsection{Deployment on live rabbits}
\label{sec:invivo}
On a single day the policy ran on four live rabbits at 5\,Hz with electrosurgery armed, to our knowledge the first time a learned policy has fired electrosurgery in vivo. Inference ran on one workstation with an NVIDIA RTX~4090 (DINOv3 in fp16) at a median 23.5\,ms per step (95th percentile 59.7\,ms), well inside the 200\,ms control period. In each animal the surgeon placed the instruments, checked the hand-over state against the demonstration band, handed control to the policy, supplied the phase by keypress and could take it back at any moment. The policy completed the appendectomy in three of the four animals (Table~\ref{tab:invivo}). In each case a single rollout carried the instruments from grasp approach to the energized cut without the surgeon touching them (Fig.~\ref{fig:invivo}(c)). In the fourth animal the left arm was handed over in a poorly conditioned inverse-kinematics branch, the joint-limit governor aborted before the policy could act, and the procedure was completed by hand. All four animals survived, and no gross abnormality was detected during one week of post-operative observation.

\begin{table}[t]
\centering
\caption{\textbf{Deployment results on four live rabbits.} The policy performed one rollout in each animal, and every value is taken from that rollout's log.}
\label{tab:invivo}
\footnotesize
\setlength{\tabcolsep}{2.5pt}
\begin{tabular}{@{}lcccc@{}}
\toprule
 & Rabbit 1 & Rabbit 2 & Rabbit 3 & Rabbit 4 \\
\midrule
Appendectomy completed & \textbf{yes} & \textbf{yes} & \textbf{yes} & \textbf{no}$^\ddagger$ \\
Policy control time (s) & 34.2 & 16.8 & 14.7 & --- \\
Cautery fired by policy (s) & 4.6 & 3.0 & 3.2 & --- \\
Peak RCM error (mm) & 0.20 & 0.21 & 0.84 & --- \\
Hand-over channels in demo band & 6/6 & 4/6 & 4/6 & --- \\
Post-operative survival (days) & 7/7 & 7/7 & 7/7 & 7/7 \\
\bottomrule
\end{tabular}
\\[2pt]
\raggedright\footnotesize $^\ddagger$Aborted at hand-over by the joint-limit governor (controller, not policy). Completed by hand.
\end{table}

\textbf{Safety.} Over all policy-controlled time that day the RCM error never exceeded 0.84\,mm against a 5\,mm abort threshold. Every sample exceeding this threshold in the day's logs occurred under human teleoperation. In every sample of cautery under policy control in the three rollouts (108 samples at 10\,Hz, 10.8\,s) the cutting jaw was closed (aperture at most 0.7\%). In the fourth animal the governor stopped the arm before any policy command reached tissue.

\section{DISCUSSION}
\label{sec:discussion}

A shaft-mounted surgical instrument-state logger with per-channel latency matching turned hand-held laparoscopic demonstrations into a bimanual policy that was selected in simulation, validated on a real rig and executed electrosurgery in live animals.

The results support each of the three contributions. First, with the surgical instrument-state logger, 533 ex-vivo episodes (74\,min) and 849 in-vivo episodes on live rabbits (135\,min) were recorded without an external camera or tracker. Per-channel latency matching was one necessary part of this pipeline, halving the dynamic depth error against the FR3 ground truth (2.69 to 1.36\,mm) and raising lift in closed loop from 19\% to 81\% of episodes. Second, closed-loop rollouts of 31 candidate configurations identified the policy design. A frozen DINOv3 backbone reduced trocar violations from 88\% to 6\% relative to ResNet-18, unfreezing its last block gave the highest full-approach rate (81\%), validation error improved as safety degraded, and the selected configuration ran on the real rig without a trocar violation. Third, retrained on 849 in-vivo demonstrations, the configuration completed the appendectomy in three of four live rabbits under shared autonomy, with the cutting jaw closed at every cautery sample and a peak RCM error of 0.84\,mm.

This study has two limitations. First, the surgeon supplied the phase in vivo, because the vision-based phase predictor agreed with the surgeon on only 24.6\% of policy control steps under the deployment endoscope, a domain shift known from phase recognition across hospitals~\cite{renz2025transfer}. Second, due to the complexity of experiments on live animals, the in-vivo study was limited to demonstrations from four rabbits and one deployment in each of four others on a single day.

\section{CONCLUSION}
Demonstrations recorded on a surgeon's own hand-held instruments were sufficient to train, select and deploy a bimanual policy that completed appendectomies with electrosurgery in three of four live rabbits.

\section*{ACKNOWLEDGMENT}
Acknowledgments of the surgeons and funding sources will be added after ICRA 2027 acceptance.

\bibliographystyle{IEEEtran}
\bibliography{refs}

\end{document}